\documentclass[preprint,12pt]{elsarticle}
\usepackage{tabularx}
\usepackage{framed} 
\usepackage{verbatim}
\usepackage{graphicx}
\usepackage[dvipsnames]{xcolor}
\usepackage{algorithm}
\usepackage{algorithmic}
\usepackage{threeparttable}
\usepackage{amsmath, amssymb}
\usepackage{breqn}
\usepackage{enumitem}
\usepackage{multirow}
\usepackage{url}
\usepackage{xurl}
\usepackage[colorlinks,citecolor=blue,urlcolor=black]{hyperref}
\usepackage{booktabs}
\usepackage{color}
\usepackage{framed}
\usepackage{subfigure}
\definecolor{shadecolor}{gray}{0.85}
\usepackage{etoolbox} 
\newtoggle{comments}
\newtoggle{anon}
\toggletrue{comments}
\toggletrue{anon}
\usepackage{todonotes}
\usepackage{makecell}
\usepackage[
n,
operators,
advantage,
sets,
adversary,
landau,
probability,
notions,	
ff,
mm,
primitives,
events,
complexity,
asymptotics,
keys]{cryptocode}
\usepackage{hhline}

\usepackage{listings}
\usepackage{xcolor}

\usepackage{enumitem}

\colorlet{punct}{red!60!black}
\definecolor{background}{HTML}{EEEEEE}
\definecolor{delim}{RGB}{20,105,176}
\colorlet{numb}{magenta!60!black}

\lstdefinelanguage{json}{
    basicstyle=\normalfont\ttfamily\scriptsize,
    showstringspaces=false,
    breaklines=true,
    frame=lines,
    backgroundcolor=\color{background},
    literate=
     *{0}{{{\color{numb}0}}}{1}
      {1}{{{\color{numb}1}}}{1}
      {2}{{{\color{numb}2}}}{1}
      {3}{{{\color{numb}3}}}{1}
      {4}{{{\color{numb}4}}}{1}
      {5}{{{\color{numb}5}}}{1}
      {6}{{{\color{numb}6}}}{1}
      {7}{{{\color{numb}7}}}{1}
      {8}{{{\color{numb}8}}}{1}
      {9}{{{\color{numb}9}}}{1}
      {:}{{{\color{punct}{:}}}}{1}
      {,}{{{\color{punct}{,}}}}{1}
      {\{}{{{\color{delim}{\{}}}}{1}
      {\}}{{{\color{delim}{\}}}}}{1}
      {[}{{{\color{delim}{[}}}}{1}
      {]}{{{\color{delim}{]}}}}{1},
}

\usepackage[utf8]{inputenc}
\usepackage{geometry}
\makeatletter
\newcommand{\linebreakand}{%
  \end{@IEEEauthorhalign}
  \hfill\mbox{}\par
  \mbox{}\hfill\begin{@IEEEauthorhalign}
}
\makeatother

\usepackage{txfonts}
\journal{Knowledge-Based Systems}
\begin{document}
\begin{frontmatter}



\title{A Systematic Framework for Tabular Data Disentanglement}


\author[ntu]{Ivan Tjuawinata}
\author[ntu]{Andre Gunawan}
\author[ntu]{Anh Quan Tran} 
\author[mastercard]{Nitish Kumar} 
\author[mastercard]{Payal Pote}
\author[mastercard]{Harsh Bansal} 
\author[ntu]{Chi-Hung Chi}
\author[ntu]{Kwok-Yan Lam}
\author[ntu]{Parventanis Murthy} 

\affiliation[ntu]{organization={Nanyang Technological University},
            country={Singapore}}
 \affiliation[mastercard]{organization={Mastercard},
             country={India}}
\begin{abstract}

 Tabular data, widely used in various applications such as industrial control systems, finance, and supply chain, often contains complex interrelationships among its attributes. Data disentanglement seeks to transform such data into latent variables with reduced interdependencies, facilitating more effective and efficient processing. Despite the extensive studies on data disentanglement over image, text, or audio data, tabular data disentanglement may require further investigation due to the more intricate attribute interactions typically found in tabular data. Moreover, due to the highly complex interrelationships, direct translation from other data domains results in suboptimal data disentanglement. Existing tabular data disentanglement methods, such as factor analysis, CT-GAN, and VAE face limitations including scalability issues, mode collapse, and poor extrapolation. In this paper, we propose the use of a framework to provide a systematic view on tabular data disentanglement  that modularizes the process into four core components: data extraction, data modeling, model analysis, and latent representation extrapolation.
We believe this work provides a deeper understanding of tabular data disentanglement and existing methods, and lays the foundation for potential future research in developing robust, efficient, and scalable data disentanglement techniques.
Finally, we demonstrate the framework’s applicability through a case study on synthetic tabular data generation, showcasing its potential in the particular downstream task of data synthesis.

\end{abstract}






\end{frontmatter}
\twocolumn
\section{Introduction}\label{sec:Intro}

Data disentanglement is a technique that aims to represent a given data as a set of more weakly related \emph{latent variables} which we call \emph{latent re-presentation}. Compared to the context of image, text, or audio data, this technique has not been as well explored in the context of tabular data.

One example where tabular data is used {\color{blue} in industries} is in industrial control systems (ICS) that manage critical infrastructures such as power
generation and distribution networks. In order to achieve its objective, ICS rely heavily on the structured  time-series data collected from
various readings of the physical system which is typically in a tabular form with rows representing time-stamped events or measurements while columns representing attributes, including temperature and device logs. In Industry 4.0, the analysis of such tabular data is essential in various applications including integrity monitoring, anomaly detection, and intrusion detection system to enable real-time decision making \cite{DX21,ABB,CE,BHMI23}. In the machine-learning-based tasks, it has been observed that learning the disentangled representation of the generative factors of the data can be useful in the completion of the machine-learning-based tasks \cite{Rid16}.

\noindent\textbf{Systematic View of Tabular Data Disentanglement.}
There have been various studies of tabular data disentanglement on different aspects based on various approaches which come with different strengths and weaknesses. This includes factor analysis \cite{Har76} with explicit but small set of possible relations, CT-GAN \cite{ctgan} that can generate conditionally generate synthetic data in a specific conditions and limitations regarding efficiency and the generated synthetic data overall quality, or VAE \cite{VAE,VAEGMM}, which considers a larger space of possible relations causing the explicitness of the relation to no longer be guaranteed while also having much more complex solution in a more advanced conditional learning. 
Furthermore, there has not been any work regarding model extrapolation where we use the trained model to predict a representation of data in a hypothetical situation. 


Such separate approaches make it difficult for the studies to help each other in providing a clearer understanding of the underlying logical flow. For instance, a universal understanding of the flow may allow us to see different approaches from solutions like factor analysis \cite{Har76}, CT-GAN \cite{ctgan} and VAE \cite{VAE, VAEGMM} as a part of a more general flow of data disentanglement.

 Secondly, as has been previously discussed, data disentanglement aims to generate a list of latent variables, 
describing the inter-relationships of the data attributes. In the effort to produce such a list, latent variables need to be chosen carefully based on the representability of the original data and the low dependency among the variables for easier analysis and reduction of redundant information, which may also cause bias in the data. 
 Since tabular data has, in general, a much more complex inter-relationships, it provides a higher need in a better understanding of tabular data disentanglement.


Such a complication may be reduced if there is a systematic way to study tabular data disentanglement. There are various further potential advantages of having a systematic view of tabular data disentanglement.

Firstly, a systematic study on tabular data disentanglement may allow us to identify its main components. This modularizes the investigation of tabular data disentanglement, allowing different approaches to be seen in a more general perspective. This allows us to differentiate between components based on the extent that they have been studied. This further allows the identification of research gaps that may exist to obtain a more complete picture of tabular data disentanglement.

Secondly, a systematic study on tabular data disentanglement involves a clearer understanding of the desired properties of such systems. This will further provide us with a way to evaluate a disentanglement system, allowing us to quantify the quality of a model and a data representation. With such quantification, there can be a unified way to evaluate the performance of a solution in a specific scenario with respect to a specific desired property. 
%

Lastly, despite the different properties that are shown in different solutions, it is not straightforward to integrate them to have a hybrid solution that takes advantage of the advances in the different approaches. If we are able to identify the main components and features of a tabular data disentanglement, we may see the existing solutions in a more systematic way, allowing a more modular way in the investigation of tabular data disentanglement. This will also allow the identification of different components of different solutions, potentially allowing different components of different solutions to be combined. This will hopefully lead to a hybrid solution possessing the features provided by each of the components while allowing limitations of any one solution to be reduced.

\noindent\textbf{Our Contributions.}

The contributions of this work can be summarized in the following.
\begin{enumerate}
    \item To provide a more comprehensive and systematic view, we have provided an exten-ded view of the tabular data disentanglement which takes more features into account. In addition, we have also included various properties desired in a tabular data disentanglement.
    \item We propose an expanded framework of ta-bular data disentanglement containing the components that are essential in the whole disentanglement system. 
    \item We analyze the tabular data disentanglement and discuss potential challenges that it introduces due to the unique features of tabular data compared to a more general setting.
   \item We provide a case study on the application of our framework under a real-world scenario.
\end{enumerate}

\section{Framework}\label{sec:Framework}

We first define the participants and components that the framework will make use of.

\subsection{Participants}\label{sec:Framework-Participants}
There are mainly two different families of participants being considered. The first family is the data owner. Here data owner(s) are those holding the data and they are responsible in processing requests and extracting information from their data. The second family is the users. Here users send a request to data owners.
\subsection{Components}\label{Sec:Framework-Components}

In this section, we discuss components of our framework.

Let $M$ be a matrix with $n$ rows and $m$ columns. For $I\subseteq \{1,\cdots, n\},$ we denote by $M_{I,:},$ the submatrix obtained from $M$ by taking all row $i$ for $i\in I$ and all the columns. Similarly, for $J\subseteq \{1,\cdots, m\},$ we denote by $M_{:,J},$ the submatrix obtained by taking all rows of $M$ with only columns corresponding to $j\in J.$ Lastly, we denote by $M_{I,J},$ the submatrix of $M$ obtained by taking rows in $I$ and columns in $J.$

For any positive integer $n,$ denote by $[n]\triangleq\{1,\cdots, n\}.$ For any set $S,$ we denote by $\mathbb{P}(S)\triangleq \{V\subseteq S\},$ the power set of $S.$
\subsubsection{Data}\label{sec:Framework-Data}
Define $m$ attribute spaces $\mathcal{A}_1,\cdots, \mathcal{A}_m.$ We note that each space can be either discrete or continuous. We further note that each $\mathcal{A}_i$ can again be multi-dimensional. In particular, in dealing with data with temporal dimension or time series data with $T$ time windows, each $\mathcal{A}_i$ is a $T$-dimensional space where each component is used to represent the data value in a different time window. Here we define the whole attribute space $\mathcal{A}=\prod_{j=1}^m \mathcal{A}_j$ and for any $J\subseteq [m],$ define $\mathcal{A}_J\triangleq\prod_{j\in J}\mathcal{A}_j.$

A tabular data record $\mathbf{x}$ is defined as a vector of length $m$ with its $i$-th entry being an element of $\mathcal{A}_i,$ that is, $\mathbf{x}\in \mathcal{A}.$ A tabular dataset $\mathcal{D}$ is defined to be a set of $n$ data records $\mathcal{D}=\left\{\mathbf{d}_1,\cdots, \mathbf{d}_n\right\}\subseteq  \mathcal{A}.$ Here, we denote the $j$-th attribute value of the $i$-th data record $\mathbf{d}_i$ as $d_{i,j}.$ 


\subsubsection{External Knowledge} Here we define an external additional information on the data that may not always be readily available from the data itself. This may include the definition of $\mathcal{A}_1,\cdots, \mathcal{A}_m,$ functional dependencies providing deterministic relation between some of the attributes, or distributions of some of the attributes and previously known latent variables. 




\subsubsection{Request}\label{sec:Framework-Request}
A request initiates the disentanglement process which is sent by a user to the data owner. Here it can be represented as a quintuple $(\mathfrak{q}_{\mathtt{Extract}},$ $\mathfrak{q}_{\mathtt{Extrapolate}},\Omega,(\alpha_R,\alpha_C),\beta)$ which will be separate-ly discussed in more detail.

\begin{enumerate}
    \item[a)] A request may contain two sets of queries, named the \emph{extraction query} and the \emph{extra-polation query}. Here the extraction query is used to extract a target data subset from the dataset held by the data owner. On the other hand, the extrapolation query is used to further modify the data representation learned from the data to represent the data under a (potentially different and hypothetical) condition. 


    Firstly, we discuss the extraction query. An extraction query $\mathfrak{q}=\left(\mathfrak{q}^{(\mathtt{c})}, \mathfrak{q}^{(\mathtt{s})}\right)$ can be divided to two parts, an \emph{attribute condition} $\mathfrak{q}^{(\mathtt{c})}$ and an \emph{attribute selection} $\mathfrak{q}^{(\mathtt{s})}.$ An attribute condition is a function $\mathfrak{q}^{(\mathtt{c})}:\mathcal{A}\rightarrow \{0,1\}$ that represents whether the input data record satisfies the condition. Given a data-set $\mathcal{D}$ containing $n$ data records and an attribute condition $\mathfrak{q}^{(\mathtt{c})},$ they induce a subset $I_{\mathfrak{q}^{(\mathtt{c})}}=\{i\in [n] :\mathfrak{q}^{(\mathtt{c})}(\mathbf{d}_i)=1\}\subseteq [n].$ Given a data record $\mathbf{d},$ we say that $\mathbf{d}$ satisfies the attribute condition $\mathfrak{q}^{(\mathtt{c})}$ if $\mathfrak{q}^{(\mathtt{c})}(\mathbf{d})=1.$

    On the other hand, an attribute selection is a set $\mathfrak{q}^{(\mathtt{s})}\subseteq [m]$ of column indices of the input data requested. Here given an extraction query $\mathfrak{q}=(\mathfrak{q}^{(\mathtt{c})},\mathfrak{q}^{(\mathtt{s})}),$ let $\mathcal{D}_{\mathfrak{q}}\triangleq \mathcal{D}_{I_{\mathfrak{q}^{(\mathtt{c})}},\mathfrak{q}^{(\mathtt{s})}},$ which we call the \emph{target window}.

    As previously described, an extrapolation query is used to further process the data representation to generate an alternative da-ta representation under the provided extrapolation condition. Here an extrapolation query again contains two parts, an \emph{attribute condition} $\mathfrak{p}^{(\mathtt{c})}$ and an \emph{attribute selection} $\mathfrak{p}^{(\mathtt{s})}.$ The extrapolation attribute selection $\mathfrak{p}^{(\mathtt{s})}$ is defined to be a set $\mathfrak{p}^{(\mathtt{s})}\subseteq \mathfrak{q}^{(\mathtt{s})}\subseteq [m].$ The extrapolation attribute condition is defined to be a distribution on the extracted attributes. Note that this includes the case of an exact value of (some) of the extracted attribute when the distribution only has one value with non-zero probability. Allowing the extrapolation attribute selection to be a distribution allows the simulation of data behaviour in diffe-rent contexts.
    
    \item[b)] An objective function $\Omega$ provides information about the intention of the requester in using the data representation. This information is important in the optimization of the disentanglement process to provide appropriate and sufficient information in the output so it performs sufficiently well in the target objective. 

    Given the queries as well as the objective function, the aim of the whole disentanglement process can be well defined. The aim of such disentanglement process is to learn an accurate representation of the \emph{true} distribution of the data satisfying the que-ries, in particular, with respect to $\Omega.$
    
    Here we consider $\Omega,$ a performance metric function that takes a data representation as an input and outputs a real number representing the performance of the representation. Without loss of generality, $\Omega$ is defined so a better performing data representation gives a higher output. The objective function can include various aspects of the process including the accuracy of a target machine learning trained and privacy guarantee in terms of the privacy leakage of original data given the output. Here we consider $\Omega$ as a pair $\Omega=(\Omega^{(\mathtt{Uti})},\Omega^{(\mathtt{Pri})})$ which corresponds to the utility and privacy objectives respectively. Assume that $\Omega^{(\mathtt{Uti})}$ measures the performance of the data representation in estimating a random variable $Z^{(\mathtt{Uti})}.$ Note that such random variable can represent both the label being trained in the target machine learning and the overall distribution of the dataset following the queries. In contrast, we may also assume that $\Omega^{(\mathtt{Pri})}$ measures the performance of a data representation with respect to the random variable $Z^{(\mathtt{Pri})},$ which corresponds to the entropy of the input dataset $\mathcal{D}.$ 
    \item[c)] \emph{Extracted Data Size:} Recall that given the input data $\mathcal{D}$ and the extraction query, we aim to generate a good representation of the true distribution of the data satisfying the query. This does not mean that the extracted data has to strictly be the one following the query. Instead, extracting more data records or attributes may help in learning the true desired distribution. Here the extracted data size is a component of the query that contains two parameters $\alpha_R,$ $\alpha_C\in \left(0,1\right)$ that provides the upper bound of the proportion of the data records and attributes that can be included in the extracted data.

    \item[d)] The last component $\beta$ provides the requirement for the size of the output. More specifically, it provides the upper bound on the number of latent variables that may be used in the disentanglement process, limiting the number of relations that can be learned and stored.
\end{enumerate}
\subsubsection{Relationship Family}\label{sec:Framework-Relationship}

Note that for a disentanglement process, we aim to learn a group of relationships the data possesses.

First, we assume the existence of a family of functions $\mathcal{F}=\{(J,f): J\subseteq [m], f:\mathcal{A}_j\rightarrow \mathtt{R}_f\}$ containing functions with domain being a direct product of some of the attribute spaces. Here for a function $f,$ we do not restrict its co-domain and instead we denote it by $\mathtt{R}_f.$ Then we define a relation to be a random variable $Z$ that corresponds to a function $(J_Z,g_Z)\in \mathcal{F}$ such that $Z=g_Z((\mathcal{A}_{j})_{j\in J_Z}).$ A relationship family $\mathcal{R}_{\mathcal{F}}$ corresponding to the family of functions $\mathcal{F}$ is the set $\mathcal{R}_{\mathcal{F}}=\{Z=g(A_J):(J,g)\in \mathcal{F}\}.$

By the distribution of $\mathcal{A}_j$ for $j\in J_Z,$ we assume that $Z$ follows the distribution with sample space $\mathtt{S}_Z$ and density function denoted by $g_Z^{(\theta_Z)}:\mathtt{S}_Z\rightarrow \mathbb{R}$ where $\theta_Z$ is the set of parameters defining the density function $g_Z^{(\theta_Z)}.$

\section{Formal Definition}\label{sec:Def}
\begin{figure*}
    \centering
    \includegraphics[width=\textwidth]{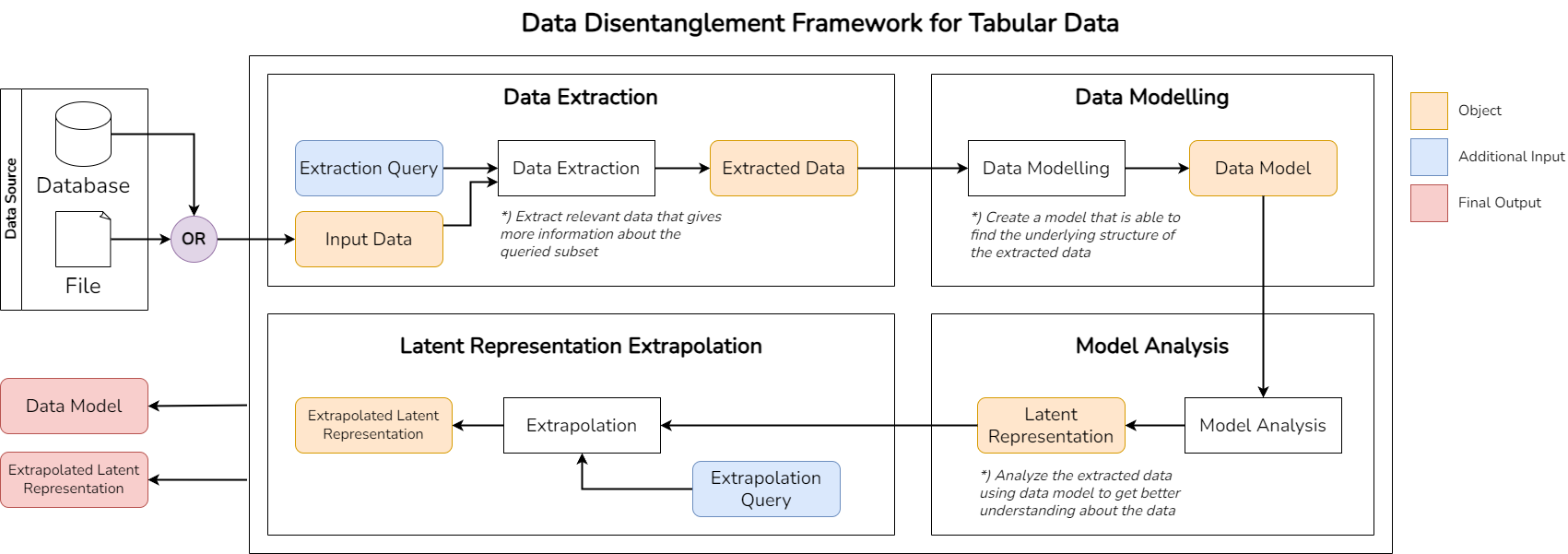}
    \caption{Proposed Framework of Tabular Data Disentanglement}\label{fig:framework}
\end{figure*}

In this section, we will provide a formal mathematical definition of data disentanglement. Such definition is based on the framework that is provided in Figure \ref{fig:framework}.

A data disentanglement process contains $5$ protocols $\Pi=(\Pi.\mathtt{Setup},\Pi.\mathtt{Extract},\Pi.\mathtt{Model}, $ $\Pi.\mathtt{Analyze},$ $\Pi.\mathtt{Extrapolate})$ where
\begin{enumerate}
    \item $\Pi.\mathtt{Setup}$ generates the necessary parameters including the relationship family based on the schema of the data, extraction query, optional extrapolation query, objective, input threshold and model size threshold.
    \item $\Pi.\mathtt{Extract}$ takes as input $\mathcal{D}, \mathfrak{q}_{\mathtt{Extract}},\Omega,$ and $\alpha$ and generates two indices sets $I^{(E)}\subseteq [n]$ and $J^{(E)}\subseteq [m].$ Based on such sets, an extracted subset of $\mathcal{D},$ denoted by $\mathcal{D}_{\mathfrak{q}_{\mathtt{Extract}}}^{(\mathtt{E})}$ $\triangleq \mathcal{D}_{I^{(E)},J^{(E)}}$ containing $m^{(E)}=|J^{(E)}|$ attributes of $n^{(E)}=|I^{(E)}|$ data records such that $n^{(E)}\leq \alpha_R n$ and $m^{(E)}\leq \alpha_C m.$ 
   
    \item $\Pi.\mathtt{Model}$ takes as inputs the following, extracted data, $\mathcal{D}_{\mathfrak{q}_{\mathtt{Extract}}}^{(E)},$ family of relationships $ \mathcal{R}_{\mathcal{F}},$ extraction query $ \mathfrak{q}_{\mathtt{Extract}},$ the objective function $\Omega,$ and output threshold $\beta.$ It outputs a set $S_{\mathcal{F}}=\{Z_1,\cdots, Z_M\}\subseteq \mathcal{R}_{\mathcal{F}}$ which we will call the data model.

    Here we assume that for each $Z_t\in S_{\mathcal{F}},$ it corresponds to $(J_t,f_t,\mathcal{R}_t,V_t)\in \mathcal{F}$ where $J_t\subseteq J^{(E)}.$ Furthermore, by the distribution of $\mathcal{A}_j$ for $j\in J_t,$ we assume that $Z_t$ follows the distribution with sample space $\mathcal{R}_t$ and the density function denoted by $f_t(X|\theta_j),$ where $\theta_j$ is the set of parameters defining the density function $f_t.$ Lastly, here we assume that $V_t\subseteq \mathbb{P}(I^{(E)})$ is a set of subsets of $\mathbb{P}(I^{(E)}).$ Here $V_t$ is used as an instruction on the different data subsets of the extracted data set that we want to estimate the distribution of $f_t$ from. Here we assume that $|V_t|=n_t.$ This allows the latent variables to be used as both common feature observable in the data or as unique distinguishing feature that can be used to distinguish between two or more groups of data. These two groups of latent variables are useful in providing a more detailed information regarding the behavior of the data.
    
    We also assume that $\Pi.\mathtt{Model}$ also outputs a pair of (possibly probabilistic) functions  $\mathtt{E}:\mathcal{A}\rightarrow \mathcal{R}_1\times\cdots\times \mathcal{R}_M$ and $\mathtt{D}:\mathcal{R}_1\times \cdots\times\mathcal{R}_M \rightarrow \mathcal{A}$ which provides the maps between the original attribute space $\mathcal{A}$ and the model space $\mathcal{R}_1\times\cdots\times \mathcal{R}_M.$ Here given a subset $X\subseteq \mathcal{A}$ and $Y\subseteq \mathcal{R}_1\times\cdots\times \mathcal{R}_M,$ we naturally extend $\mathtt{E}$ with input $X$ to be the set obtained by running $\mathtt{E}$ to each element of $X.$ We similarly extend $\mathtt{D}$ for an input set $Y.$ 

    \item $\Pi.\mathtt{Analyze}$ takes as input the data model $S_{\mathcal{F}}$ and the extracted data $\mathcal{D}_{\mathfrak{q}_{\mathtt{Extract}}}^{(E)}$ and outputs the extracted data representation 
    \[\mathcal{R}_{\mathfrak{q}_{\mathtt{Extract}}}^{(E)}\triangleq\left\{
    \begin{array}{c}
         (Z_1,(\hat{\theta}^{(1)},\cdots, \hat{\theta}^{(1)}_{n_1})),  \\
         \cdots,\\
         (Z_M,(\hat{\theta}^{(M)}_1,\cdots, \hat{\theta}^{(M)}_{n_M}))
    \end{array}
    \right\}\] where $\hat{\theta}^{(t)}_\ell$ is an estimate of the parameter $\theta_t$ used in $f_t$ based on the $\ell$-th subset in $V_t$ of the extracted training data $\mathcal{D}_{\mathfrak{q}}^{(E)}.$ The objective of this sub-protocol is to learn the distribution for the latent variables $Z_i$ with respect to the extracted data $\mathcal{D}_{\mathfrak{q}_{\mathtt{Extract}}}^{(E)}.$ Here it may also include other analysis such as the analysis of the relation as well as the features. Here we say that a data representation $\mathcal{R}$ is compatible with the data model $\mathcal{S}$ if the $i$-th component of $\mathcal{R}$ contains a valid distribution and parameter for the $i$-th random variable in $\mathcal{S}$ for all $i.$

    \item $\Pi.\mathtt{Extrapolate}$ takes as input a data model $S_{\mathcal{F}},$ an extracted dataset $\mathcal{D}_{\mathfrak{q}_{\mathtt{Extract}}}^{(E)},$ a data representation $\mathcal{R}_{\mathfrak{q}_{\mathtt{Extract}}}^{(E)}$ of the extracted data-set $\mathcal{D}_{\mathfrak{q}_{\mathtt{Extract}}}^{(E)},$ and an extrapolation query\\ $\mathfrak{q}_{\mathtt{Extrapolate}}$ and outputs a new data representation $\mathcal{R}_{\mathtt{Extrapolate}}.$ In this sub-protocol, given the learned data model and representation, we estimate a new data representation which represents the dataset under the extrapolated condition based on the extrapolation query. In the following, we consider three different ways the extrapolation protocol $\Pi.\mathtt{Extrapolate}$ can be used. 
    
    Firstly, we consider some notation regarding the attribute space. We consider an attribute space $\mathcal{A}$ and we assume that $\mathcal{A}$ contains the set of values of such an attribute can take. We note that in the real life application, the value of $\mathcal{A}$ may be unknown to anyone. This can be in the form of an unknown category in a categorical attribute or a value that is beyond the observed range in a continuous attribute.
    
    We first illustrate our definition by considering an $m$-dimensional real-valued attribute. Suppose that based on the available data, we have observed values $\mathbf{x}_1,\cdots, $ $\mathbf{x}_n\in \mathbb{R}^m.$ This provides us with a finite set of observed data $S^{(\mathtt{O})}=\{\mathbf{x}_1,\cdots, \mathbf{x}_n\}.$ 
    
    Furthermore, for $j=1,\cdots,m$ define $\mathcal{A}^{(j)}$ $=\{(\mathbf{x}_1)_j,(\mathbf{x}_2)_j,\cdots, (\mathbf{x}_n)_j\}\subseteq \mathbb{R}.$ Let $a^{(j)}_1\leq\cdots\leq a^{(j)}_{n_j}\in \mathbb{R}$ such that $\mathcal{A}^{(j)}=\{a^{(j)}_1,\cdots, a^{(j)}_{n_j}\}.$ Using $\mathcal{A}^{(1)},\cdots, \mathcal{A}^{(m)},$ we may extend $S^{(\mathtt{O})}$ to contain more elements of $\mathcal{A}.$ More speci-fically, define the grid $S^{(\mathcal{G})}\triangleq \prod_{i=1}^{m}\mathcal{A}^{(j)}\cap \mathcal{A}.$ Note that $S^{(\mathcal{G})}$ is still a finite set containing $S^{(\mathcal{O})}$ while potentially allowing more values to be considered. Define $S^{(1)}\triangleq S^{(\mathcal{G})}\setminus S^{(\mathtt{O})},$ the set of elements in $S^{(\mathcal{G})}$ that were not observed.
    
    Next, for $j=1,\cdots, m,$  we define further $\mathcal{I}^{(j)}=[a^{(j)}_1,a^{(j)}_{n_j}]$ the smallest interval containing $\mathcal{A}^{(j)}.$ This allows us to define a cuboid $S^{(\mathtt{C})}\triangleq \prod_{i=1}^m \mathcal{I}^{(j)}\cap \mathcal{A}.$ This extension considers the smallest cuboid containing $S^{(\mathtt{O})}.$ Here intuitively, because the $j$-th entry of such attribute takes values $a^{(j)}_1$ and $a^{(j)}_{n_j},$ combined with the fact that such entry is a continuous variable, it \emph{implies} that such entry may take any value between $a^{(j)}_1$ and $a^{(j)}_{n_j}.$ Define the second and larger extension set $S^{(2)}\triangleq S^{\mathtt{(C)}}\setminus S^{(\mathtt{G})}$ that contains all unobserved values in $S^{(\mathtt{C})}$ that do not belong to the grid $S^{(\mathcal{G})}.$ Lastly, we define the third and largest extension set $S^{(3)}\triangleq \mathcal{A}\setminus S^{(\mathtt{C})},$ containing all values whose at least one of its entries is not implied by the observed data.

    It is easy to see that the definitions can be easily extended to the entire attribute space when all the attributes are continuous. Furthermore, the concepts of the observed set and the first extension set can be easily extended to the case when at least one entries of the attribute is categorical. Furthermore, once the implied set of a categorical attribute can be defined, we can also extend the definition of the second and third extension sets. For instance, such a concept may be easily defined if the set of categories is partially ordered. Denote its partial order relation by $\preccurlyeq.$ In such a case, given the observed values $\mathcal{A}^{(j)}\subseteq \mathcal{A}_j,$ its implied set is then defined as
     $\mathcal{I}^{(j)}\triangleq \{x\in \mathcal{A}_j:\exists a,b\in \mathcal{A}^{(j)}, a\preccurlyeq x, x\preccurlyeq b\}.$ Note that depending on the context, the implied set for a given categorical attribute can be defined in a different way.

     Extrapolation query can then be divided to three possible forms depending on the support of the distribution provided in the extrapolation attribute condition. Here the first two can be seen more as interpolation queries. The first level of interpolation is when the support of the requested distribution is a subset of $\mathcal{A}^{(\mathtt{G})}$ with a non-empty intersection with $S^{(1)}.$ It is easy to see that although the support may contain a data record that has not been observed from the given data, each of its attribute values has appeared in at least one of the observed data.

     The second level of interpolation extends the support for the requested distribution to be a subset of $\mathcal{A}^{(\mathtt{C})}$ with a non-empty intersection with $S^{(2)}.$ In this case, we may no longer have the guarantee that a data record in the support must have each of its attribute values to appear in at least one of the observed data. However, we still have the guarantee that such value is still implied to exist from the observed data which may still provide some information on how the data record should behave.

     The last form is the extrapolation where the support can now be any element of $\mathcal{A}$ with a non-empty intersection with $S^{(3)}.$ In this case, we note that since a data record in the support may have an attribute value that is not implied to exist by the observed data, it has the least amount of information that may be learned from the provided data record.
    
\end{enumerate}

We further define some properties of $\Pi$ which may be essential in different settings. Here we provide brief intuition behind each property while formal definition of such properties can be found in Supplemental Material. 

\begin{enumerate}
    \item \emph{Correctness}: A data disentanglement protocol $\Pi$ is said to be correct if each property returns the expected output. More speci-fically, data extraction provides a subset of the input data with dimension within the input storage limit. Data modelling outputs a set of random variables of size at most the output storage limit while model analysis provides a distribution for each random variable that the extracted data follows. Here if the data model obtained from the data modelling protocol is of size at most $\beta,$ we also say that the model is \emph{$\beta$-compact}. Lastly, the output of the extrapolation protocol is a set of distributions that is compatible with the corresponding random variables in the data model.
    

      \item \emph{Covering:} For a given similarity function that measures the likelihood of a data record to satisfy the input condition and $\tau\in(0,1),$ an extracted dataset is said to be $\tau$-covering if any included data record has a probability of at least $\tau$ to satisfy the input condition.
      
      \item \emph{Optimality of Disentanglement and Representation:} We evaluate the result of a data modelling by measuring the conditional entropy of the required information conditioned on the set of random variables in the data model. This includes the objec-tive-based optimality where we consider the conditional entropy of a set of target random variables given the data model and data-based optimality where we consider the conditional entropy of the joint distribution of the attributes in the data given the data model. In addition to the uti-lity objective which provides a minimalization function of the entropy of the target information, it also includes the privacy objective which provides a maximization aspect in the optimization problem where we want to maximize the entropy of the protected random variable.
     
     \item \emph{Independence}: Another objective of disentanglement is to have the set of random variables in the data model to have weaker correlation compared to the attributes in the extracted data. Hence, we may also evaluate the resulting data model through an independence metric of the data model. We note that such independence metric can also be combined with the optimality property discussed before to have an optimization function that optimizes the random variables from both the optimality and the independence of the representation, which is an approach that is considered, for example, in \cite{VAE}.
    
    \item \emph{Extrapolation Accuracy:} Lastly, since our scheme contains the extrapolation protocol, a property that measures the quality of the extrapolation is necessary. Here we may measure the accuracy of the extrapolation method with respect to some statistical distance function by measuring such distance between the generated representation and the representation of data generated through the true distribution of such hypothetical situation.
    
\end{enumerate}
\section{Discussion on Components of Data Disentanglement}\label{sec:Discussion}
 In this section, we briefly discuss the four main components of a data disentanglement process, namely data extraction, data model, model analysis, and model extrapolation. In each component, we will discuss its main objective, its importance, why it is inseparable from the disentanglement process, existing work on such component, as well as potential challenges that may be interesting to be investigated further. A more detailed discussion can be found in the Supplemental Material.
\subsection{Data Extraction}
 Data extraction is a protocol that takes a data source and a request to extract a subset of data that will be the input for the data modelling process. Generally, data disentanglement process assumes that an input data is provided. However, in the context of tabular data, the data to be processed may be a part of larger data source that we may not want to process as a whole since not all the data may be relevant to our objective and it may even be infeasible to do so. Here, given the subset of the data source that satisfies the conditions in the request, which we call \emph{target window}, it attempts to enrich it by including more data outside of the target window to provide a more complete information regarding the \emph{true distribution} of the target window. This is important to allow the subsequent processes to model, represent, and analyze the target window better. However, this also shows the importance of the subsequent processes to influence the data extraction process to ensure that data extraction is done with respect to the need from the other processes.
Although there have been various works attempting to investigate similar problems, such a problem is in general less explored. Some works investigated \emph{data prunning}, whose objective is to prune large data source to smaller one that represents the original data \cite{dsdm,lessismore} and  \emph{data distillation}, whose objective is to generate a smaller synthetic data to represent the larger input data 
\cite{datasetdistillation,distributionmatching,matchingtrainingtrajectories}.
The capability of a data set to represent another data set can be defined based on various metrics, including the similarity of the performance of predictive models trained using the two data sets, the statistical distance between the two data sets \cite{distributionmatching}, and the differences in parameters between the models trained using the two data sets \cite{matchingtrainingtrajectories}.
Such works have shown to be essential in alleviating the learning cost, especially when the original data is infeasibly large. 
On the other hand, there have also been some works that investigated the problem of data enrichment which enlarges the provided data while maintaining the data distribution \cite{smote, generativemtd}. Such works have shown to be important in the case of imbalanced data to reduce the effect of imbalance by enlarging the smaller groups without affecting its distribution. Although the two directions are important to investigate, there are some differences on the objective compared to the data extraction in our proposed framework. More specifically, data extraction aims to enrich the target window by including data outside the target window that provides more information regarding the true distribution of the target window. So here, in contrast to the data distillation, we aim to include more data in the larger input data instead of distilling the information of the target window. On the other hand, the focus is also different from the data enrichment study. Here, instead of including data that supports the sample distribution of the target window, we aim to predict its true distribution, which may not be the same as the one we observe from the target window. This shows that, although the works in the other directions may provide important insights in how data extraction can be achieved, they may not be able to fully capture its objective, showing a potential research gap between existing studies and the objective that we want to achieve in data extraction.
\subsection{Data Modelling}
Data modelling is a core process in data disentanglement which aims to disentangle the inter-relationships of the attribtues in the data provided to such process. Provided with a predefined relationship family and a target random variable, the result of such a process is a set of variables from such family which we call \emph{latent variables} to be used to represent the target random variable. The aim of data modelling is to have such a set of latent variables, which we call \emph{data model} to be more informative and easier to process for the subsequent processes or tasks following the given objective. The simplest way to determine such variables are to deterministically and explicitly define the variables. On one hand, this provides clear latent variables making analysis of such model much simpler. However, due to its deterministic nature, it may only capture a small space of possible latent variables.An example of such approach is the factor analysis \cite{factoranalysistutorial}.
A more common way in generating such set is through loss function based optimization where the loss function is defined based on the objectives imposed to the process \cite{VAE, ctgan}. There are in general two main objectives when constructing such set. The first main objective is to have such a set to contain sufficient information on the target random variable to be able to generate an accurate representation of the target random variable in the \emph{latent space}, the space generated by the latent variables, providing a more complete understanding of both the data and the target random variable. Here in the works on data modelling, we have in general two or more types of latent variables 
\cite{subtab, switchtab}. 
Firstly, a latent variable, which we call \emph{common feature}, may provide a general description of the data behavior that is present throughout the data. Alternatively, a latent variable, which we call \emph{unique feature}, provides a distinguishing behavior separating subsets of the data. Such features provide more detailed variations of the data.
The second main objective of the data modelling module, as the term ``disentanglement'' suggests, is to weaken the statistical dependence between variables. When variables have strong statistical dependence, identifying its distribution requires the analysis of the joint behavior, causing such effort to be more complex. Consequently, processing a set of strongly dependent variables is also much more complex than processing a set of more weakly dependent variables. Because of this, ideally, a statistically independent set of latent variables is desired. However, in general, it may not be feasible to obtain such statistical independence. Instead, existing works \cite{factorvae} aim to construct latent variables whose statistical dependence to each other is weaker than that of the attributes in the original data. Due to the relaxed requirement, there needs to be investigation on the remaining statistical dependence. On one hand, some works  consider methods such as \emph{stacking} \cite{HQVAE, NVAE}, statistical analysis of the dependence \cite{CausalVAE}, or simply assuming its non-existence when it is sufficiently weak.
 Potential research gaps include large-space explicitly defined latent variables, conditional modelling, and the generalization to multi-table or temporal-based tabular data.
\subsection{Model Analysis}
Given the data model from the previous module, model analysis is a module that further studies the input data in the perspective of the data model with objective of learning its distribution and properties. Such study enables a more complete understanding of the data behavior in the perspective that has been determined to be important in the previous module. Due to the close relation between the objectives of this module and the previous module, it is clear that the two modules cannot be done separately. 

In contrast to other types of data, tabular data has more complex behavior to learn. Works on image data disentanglement generally assume that the latent variables are normally distributed \cite{VAE}. Although this may be sufficiently general in the context of image, it is not as apparent when dealing with tabular data where more complex distribution may be observed. Some works have considered more complex form of distributions such as mixture of multiple normal distributions \cite{VAEGMM,GMVAE,tGMVAE}. Alternatively, instead of aiming to identify the original distribution of the latent variables, there have also been works that aim to construct a transformation function that transforms the original random variable to standard normal \cite{tabsyn}. In either case, it is quite apparent that more careful studies on the distribution of tabular data model are required.

\subsection{Latent Representation Extrapolation}

Previous three modules learn the distribution and properties of the original data is the main objective. In contrast, extrapolation aims to provide additional value to the learned representation. Extrapolation allows the generation of a new representation of the model based on a set of control variables we call \emph{condition variables} to enhance the usability of the learned model and representation of the extracted data. Here, given a set of conditions that can be represented as the distribution of some variables, the aim is to generate a new representation in the perspective of the data model that is predicted to occur when the provided conditions are present. Considering that the extrapolation can be seen as the final objective of the entire data disentanglement process, the other modules will need to be performed with the extrapolation objective considered, making such a module inseparable from the other modules.

In the context of image data, conditional generation of representation and synthetic data under the given conditions have been studied \cite{ccvae,dptvae}. Another work has also used conditional data disentanglement for cell analysis \cite{cvaecell}. However, direct translation of such work to tabular data is not trivial. Tabular data, which has different types of attributes needs to be handled carefully to avoid invalid result. Furthermore, in contrast to image data that has a more visual nature, assessment of the extrapolation also requires further studies.

\section{Data Disentanglement Use Case of Synthetic Data Generation}\label{sec:DataSynthesis}

In this section, we provide a case study on the application of tabular data disentanglement for data synthesis. We  discuss the difference between a standard data disentanglement-based data synthesis approach and the differences that we can observe when approaching it through the systematic view that we have considered. 
The application of tabular data disentanglement through data synthesis can be seen in, for instance the introduction of fraud pattern from one country to the fraud detection system in another country or the modelling of a spending behaviour in a certain weather condition from one country given the analogous data in another. 

First, we discuss a generic approach that is followed by existing works, such as Tabsyn \cite{tabsyn}, Switchtab \cite{switchtab}, VAEGMM \cite{VAEGMM}, and TVAE \cite{ctgan}. Here given an input data, it is processed using data disentanglement methods such as VAE with the objective of generating a latent representation that is capable of recovering the original data while having some independence guarantee through various metrics such as KL-Divergence. In order to learn the distribution of such representation, various methods may be used along with the disentanglement method, including the fitting to a Gaussian Mixture Model to allow the representation to follow a more sophisticated distribution \cite{VAEGMM} or the training and generation of a transformation function between the latent representation and a set of independently and standard normally distributed random variables \cite{tabsyn}. Having this latent representation and its distribution, the solutions proceed to generate a synthetic latent representation following the estima-ted distribution or through the use of the  transformation function, which is then decoded using the decoder generated during the data disentanglement method to transform the data from the latent space to the original space. Such approach may be equipped with the capability of transforming the representation to a scenario defined by the conditions provided \cite{ctgan,dptvae}. In such case, intuitively, this is done by first fixing the attributes related to the condition, then have the learning process to learn the conditional distribution of the data given the value of the condition attributes. Then having this, when a new condition is given regarding such attributes, where, without loss of generality, we can assume that the condition is in the form of the distribution of such attribute, such solutions use the condition as the prior to generate the data representation and its distribution based on the provided prior. Having this new data representation and its distribution, synthetic data generation can then be performed in a similar way as previously discussed.

Next we briefly discuss various sectors that may require further investigation on the generic data synthesis process when considering the systematic view previous discussed. Firstly, in the real-life application, data is often not in the state that is ready for the data disentanglement process. For instance, it may be a part of a much larger dataset where processing the entire dataset may not be feasible. In such case, investigation on how such process can be done with the subsequent tasks in mind is important to ensure that the appropriate data subset is extracted for our objective. To the best of our knowledge, such data processing has not been extensively studied and existing studies such as \cite{datasetdistillation,neuralscalingdatapruning} are designed for a slightly different objective which is independent of the subsequent tasks. 

Furthermore, in such works, in order to evaluate the quality of the resulting synthetic data, various evaluation metrics have been proposed, for instance \cite{matchingtrainingtrajectories,distributionmatching}. However, we note that in such works, evaluation step is often independent of the training step and it does not affect how the training is done. In the definition that we have considered, such evaluation metrics can be seen as a part of the request, which, when being used as an input for the other protocols, allows a more focused training phase. On the other direction, such evaluation metric should also be defined based on the properties and features that are provided in such request. This allows the evaluation metric to measure the quality precisely based on the objective that we would like to achieve on the synthetic data. For instance, when synthetic data is being used to represent the original data, evaluation through ML efficacy may have less relevance. Instead, we may consider the optimality of disentanglement and representation to better evaluate the synthetic data based on the given objective. This discussion also provides us with insights that conditional disentanglement and conditional generation have shown to be more essential where training and generation are performed based on a specific conditions rather than having it to be done in a more general direction.

Lastly, we note that in general, existing solutions have the disentanglement method, conditional generation, and evaluation to be done independently. When disentanglement is not done conditioned on the desired objective and extrapolation scenario, the components may work in different objectives, potentially harming the performance of the overall process. Here, we can see that allowing disentanglement protocols to be controlled by the request, all the components can work with one common objective in mind. Ideally, this may provide us with a more focused process towards the defined direction. However, this also provides us with some challenges. First-ly, as has been previously discussed, this suggests the importance of having disentanglement method to be conditioned on some objective. Secondly, as opposed to only have the condition to be based on the value of some attributes, as has been done in works such as \cite{ctgan,dptvae}, conditions can be made more general to provide a more versatile condition. For instance, the condition can be based on the country of origin. Here such condition may not only control a specific attribute. Instead, it may provide some requirement on the distribution of various attributes that are implied by such country of origin. Thirdly, as opposed to the previous argument that components are not sufficiently related, it also causes some potential questions regarding the close relation between different components. Here, we consider the approach used by VAE-based solution. Here we note that without the enhancement techniques such as \cite{VAEGMM} and \cite{tabsyn}, the output of the VAE approaches provide both the set of latent variables and its distribution. This can be observed by the loss function in the original VAE, which has the KL divergence (relative to a set of independently and normally distributed random variables term) to impose the independence condition. Here due to this term, it forces the data to follow the normal distribution while disregarding potential relation between the variables despite the possibility of such relation to contain important information regarding the original data. In contrast, if the data modelling and the model analysis steps in VAE can be done in a more separate manner, such imposement may not be necessary. This may allow the resulting latent representation to better represent the data since it allows such representation to have a more general distribution that may not be captured by the normal distribution.

\section{Conclusion and Future Work}\label{sec:conc}
As has been discussed, tabular data and its analysis, including approaches based on disentanglement, have shown to be very important, especially in the real-life application. Due to its importance, we believe that a systematic view of tabular data disentanglement is essential.

In this work, we have presented our proposed framework which identified four main components which we believe to be essential in its applicability in various real-life applications. Despite them being more extensively studied in different setting with, for instance, image data, data modelling and model analysis are less explored in the context of tabular data. Although techniques and ideas from other contexts may be used in the tabular data context, studies on approaches dedicated to tabular data may be helpful in providing insights on the characteristics of tabular data that may not be observable in other context. On the other hand, data extraction and model extrapolation are less explored. 

As has been previously discussed, data extraction and representation extrapolation are two essential components of data disentanglement. Despite their importance, there have been less extensive works on them, even in the context of image data. This clearly shows a potential gap 
in our understanding of data disentanglement in general.


Lastly, although investigation of disentanglement of tabular data in the form of a single table is important, it should be seen as a starting point towards a more general form of tabular data which are more commonly used in real life. This includes generalization towards multi-table setting or the introduction of temporal dimension in the tabular data. In either case, we note that our current framework is still applicable although further investigation is needed. We believe that this work can provide a starting point and insights towards such generalizations.

%

\bibliographystyle{plain}
\bibliography{sec/reference}
\onecolumn
\appendix
\section{Formal Definitions of Properties of Data Disentanglement Process}\label{sec:FormalDefProps}
We first simplify some notations. Let $\mathcal{D}\in \mathbb{P}(\mathcal{A})$ be any input data, $\mathfrak{q}^{(in)}=(\mathfrak{q}^{(in,c)},\mathfrak{q}^{(in,s)})$ be an extraction query and $\mathfrak{q}^{(out)}$ be an extrapolation query, $\Omega$ be an objective function, $(\alpha_R,\alpha_C)$ be an input size limit, $\beta$ be an output size limit, and $\mathcal{R}_{\mathcal{F}}$ be a relationship family. 

We further let $\mathcal{D}^{(E)}=\mathcal{D}_{I^{(E)},J^{(E)}}$ be the extracted data set generated by $\Pi.\mathtt{Extract}$ with extraction query $\mathfrak{q}^{(in)},$ objective $\Omega,$ and input size limit $(\alpha_R,\alpha_C).$ We further let $\mathcal{S}_{\mathcal{F}}$ be the model obtained from $\Pi.\mathtt{Model},~ \mathcal{R}^{(E)}$ be the output of $\Pi.\mathtt{Analyze}$ under model $\mathcal{S}_{\mathcal{F}}$ and extracted data $\mathcal{D}^{(E)}$ while $\mathcal{R}^{(out)}$ be the output of $\Pi.\mathtt{Extrapolate}$ with the extrapolation query $\mathfrak{q}^{(out)}.$

\begin{enumerate}
    \item \emph{Correctness}: A data disentanglement protocol $\Pi$ is said to be correct if the following holds
    \begin{itemize}
        \item \emph{Correctness of $\Pi.\mathtt{Extract}$:} $I^{(E)}\subseteq[n], J^{(E)}\subseteq [m]$ with $|I^{(E)}|\leq \alpha n$ and $|J^{(E)}|\leq \alpha m.$
        \item \emph{Correctness of $\Pi.\mathtt{Model}$:} $\mathcal{S}_{\mathcal{F}}\subseteq \mathcal{R}_f$ containing at most $\beta$ relations. Furthermore, for any valid data $\mathcal{D}\subseteq \mathcal{A}$ and the output $(\mathcal{S}_{\mathcal{F}},\mathtt{E},\mathtt{D})$ of $\Pi.\mathtt{Model},\mathtt{D}(\mathcal{S}_{\mathcal{F}})$ is a valid dataset.
        \item \emph{Correctness of $\Pi.\mathtt{Analyze}$:}
        For any $(Z,\hat{\theta})\in \mathcal{R}^{(E)}$ with its corresponding function $f_Z,$ domain $J_Z,$ sample space $\mathtt{S}_Z$ and density function $g_Z^{\theta},$ the sample data $\{z_i = f_Z((\mathbf{d}_i)_{J_Z}):i\in I^{(E)}\}$ follows the density function $g_Z^{\theta}$ where $\mathbb{E}(\theta|\mathcal{D}^{(E)})=\hat{\theta}.$ Here the operator $\mathbb{E}$ denotes the expectation operation.
        \item \emph{Correctness of $\Pi.\mathtt{Extrapolate}$:}
        For any valid data $\mathcal{D}\subseteq \mathcal{A},$ the output $(\mathcal{S}_{\mathcal{F}},\mathtt{E},\mathtt{D})$ of $\Pi.\mathtt{Model},$ the output $\mathcal{R}_{\mathfrak{q}_{\mathtt{Extract}}}^{(E)}$ of $\Pi.\mathtt{Analyze},$ and $\mathcal{R}_{\mathtt{Extrapolate}}$ the extrapolated data representation, we have that $\mathcal{R}_{\mathtt{Extrapolate}}$ is compatible with $\mathcal{S}_{\mathcal{F}}$ and $\mathtt{D}(\mathcal{R}_{\mathtt{Extrapolate}})$ is a valid dataset.
        
     \end{itemize}
     \item \emph{Model Compactness:} For $B\in \mathbb{Z},$ we say that a model $\mathcal{S}_{\mathcal{F}}$ is $B$-compact if $|\mathcal{S}_{\mathcal{F}}|\leq B.$
\item \emph{Covering:} Given $J^{(E)}$ and $\mathfrak{q}^{(c)},$ let $\mathfrak{c}:\mathcal{A}\rightarrow [0,1]$ be a function that measures the similarity of a data record to those satisfying the input attribute condition. Here $\mathfrak{c}(\mathbf{d})=1$ if and only if $\mathfrak{q}^{(c)}(\mathbf{d})=1.$  For $\tau\in(0,1),$ we say that $I^{(E)}$ is $\tau$-covering with respect to the classifer $\mathfrak{c}$ if for any $i\in I^{(E)},\mathfrak{c}(\mathbf{d}_i)>\tau.$ Here note that $\mathfrak{c}$ can be chosen to be a model that provides the probability that the given data record satisfies the attribute condition. 
      
      \item \emph{Objective-based Optimality of Disentanglement:} Before we discuss about the optimality definition, we first define the valid disentanglement output pair. A pair $(J,\mathcal{S})$ is said to be a valid disentanglement output pair with respect to the compactness parameter $\beta$ and attribute number limit $\alpha_C$ if $J\subseteq[m],|J|\leq \alpha_Cm,\mathcal{S}\subseteq \mathcal{R}_{\mathcal{F}}, |\mathcal{S}|\leq \beta,$ and for any $Z\in \mathcal{S}_{\mathcal{F}}$ with the respective domain index $J_Z,$ we have $J_Z\subseteq J.$ We define the set $\mathtt{S}^{(\mathtt{dis},\gamma,\alpha_C)}$ to be the set of all valid disentanglement output pair.
      
      We say that a valid disentanglement output pair $(J^{(E)},\mathcal{S}_{\mathcal{F}})$ is an \emph{optimal disentanglement output} with respect to the compactness parameter $\beta,$ attribute number limit $\alpha_C,$ and utility random variable $Z^{(\mathtt{Uti})}$ if it provides the \emph{minimum} value in the following optimization problem
      \[ \min_{(J^{(E)},\mathcal{S}_{\mathcal{F}})\in\mathtt{S}^{(\mathtt{dis},\gamma,\alpha_C)} }
      \mathbb{H}(Z^{(\mathtt{Uti})}|\mathcal{S}_{\mathcal{F}}).\]
      Here $\mathbb{H}$ represents the entropy function.
      Let $(J^\ast, \mathcal{S}^\ast)$ be an optimal disentanglement output. Then we say that $\Pi.\mathtt{Extract}$ and $\Pi.\mathtt{Model}$ produce an $\epsilon$ optimal disentanglement output if the resulting $(J^{(E)},\mathcal{S}_{\mathcal{F}})$ satisfies 
      \[\mathbb{H}(Z^{(\mathtt{Uti})}|\mathcal{S}_{\mathcal{F}})<\mathbb{H}(Z^{(\mathtt{Uti})}|\mathcal{S}^\ast)+\epsilon.\]

      \item \emph{Objective-based Optimality of Representation:} We note that the inclusion of a data record to $\mathcal{D}^{(E)}$ based on the derived $I^{(E)}$ affects the privacy objective represented by $Z^{(\mathtt{Pri})}$ through the resulting data representation $\mathcal{R}^{(E)}.$ As before, we first define the notion of valid representation pair. A pair $(I,\mathcal{R})$ is said to be a \emph{valid representation pair} with record number limit $\alpha_R$ and corresponding valid disentanglement output pair $(J,\mathcal{S})\in \mathtt{S}^{(\mathtt{dis},\gamma,\alpha_C)}$ if $I\subseteq [n], |I|\leq \alpha_R n$ and $\mathcal{R}$ is the output of $\Pi.\mathtt{Embed}$ given $\mathcal{D}_{I,J}$ as the extracted data and $\mathcal{S}$ as the disentangled model. We define the set $\mathtt{S}^{(\mathtt{emb},\alpha_R,J,\mathcal{S})}$ to be the set of all such valid representation pairs. 
      
      A valid representation pair $(I,\mathcal{R})$ is said to be an \emph{optimal data record selection} with respect to the covering parameter $\tau,$ record number limit $\alpha_R,$ valid disentanglement output pair $(J,\mathcal{S})$ and privacy random variable $Z^{(\mathtt{Pri})}$ if it provides the \emph{maximum} value in the following optimization problem
      \[\max_{(I,\mathcal{R})\in \mathtt{S}^{(\mathtt{emb},\alpha_R,J,\mathcal{S})}}
      \mathbb{H}(Z^{(\mathtt{Pri})}|\mathcal{R}^{(E)}).\]
      Analogous to the optimality of disentanglement output, letting $(I^\ast,\mathcal{R}^\ast)$ be an optimal data record selection output, we say that $\Pi.\mathtt{Extract}$ and $\Pi.\mathtt{Analyze}$ produce an $\epsilon'-$optimal data record selection output if the resulting $(I^{(E)},\mathcal{R}^{(E)})$ satisfies
      \[\mathbb{H}(Z^{(\mathtt{Pri})}|\mathcal{R}^{(E)})> \mathbb{H}(Z^{(\mathtt{Pri})}|\mathcal{R}^\ast)-\epsilon'.\]

      Note that the two optimizations which involve $\Pi.\mathtt{Extract}, \Pi.\mathtt{Model},$ and $\Pi.\mathtt{Analyze}$ may not be done separately. Instead, there may only be one optimizations to produce the quadruple $\mathbf{v}\triangleq (I^{(E)},J^{(E)},\mathcal{S}_{\mathcal{F}},\mathcal{R}^{(E)}).$ Here we define say a quadruple $\mathbf{v}=(I,E,\mathcal{S},\mathcal{R})$ is a valid representation model if $(J,\mathcal{S})\in \mathtt{S}^{(\mathtt{dis},\gamma,\alpha_C)}$ and $(I,\mathcal{R})\in \mathtt{S}^{(\mathtt{emb},\alpha_R,J,\mathcal{S})}.$ We define the set of all valid representation model quadruple to be $\mathtt{S}^{(\gamma,\alpha_R,\alpha_C)}.$

      We further assume that the two entropy functions to be optimized will be combined through some performance metric function $\varphi:\mathbb{R}^2\rightarrow\mathbb{R}$ where we assume it to be non-decreasing with respect to $\mathbb{H}(Z^{(\mathtt{Pri})}|\mathcal{R})$ and non-increasing with respect to $\mathbb{H}(Z^{(\mathtt{Uti})}|\mathcal{S}).$  

      We can then define a valid representation model $\mathbf{v}$ to be an optimal representation model with respect to the objective variables $Z^{(\mathtt{Pri})}$ and $Z^{(\mathtt{Uti})}$ with combining function $\varphi$ if it provides the \emph{maximum} value in the following optimization problem

      \[\max_{\mathbf{v}\in \mathtt{S}^{(\gamma,\alpha_R,\alpha_C)}}\varphi(\mathbb{H}(Z^{(\mathtt{Uti})}|\mathcal{S}),\mathbb{H}(Z^{(\mathtt{Pri})}|\mathcal{R})).\]

      Given that the optimal value is achieved by a valid quadruple $\mathbf{v}^\ast,$ we say that $\mathbf{v}$ is $\epsilon$-optimal if 
      $\varphi(\mathbb{H}(Z^{(\mathtt{Uti})}|\mathcal{S}),\mathbb{H}(Z^{(\mathtt{Pri})}|\mathcal{R}))>\varphi(\mathbb{H}(Z^{(\mathtt{Uti})}|\mathcal{S}^\ast),\mathbb{H}(Z^{(\mathtt{Pri})}|\mathcal{R}^\ast))-\epsilon
      .$
      
      We note that the simplest way to define such function $\varphi$ is to have it as a weighted sum of the two entropy functions. Without loss of generality, we assume that the weight of the utility-related entropy function to be $\lambda^{(\mathtt{Uti})}\in \mathbb{R}_{>0}$ while the weight of the privacy-related entropy function to be $1.$ Hence we define $\varphi(\mathbb{H}(Z^{(\mathtt{Uti})}|\mathcal{S}),\mathbb{H}(Z^{(\mathtt{Pri})}|\mathcal{R})) =\mathbb{H}(Z^{(\mathtt{Pri})}|\mathcal{R}) - \lambda^{(\mathtt{Uti})} \mathbb{H}(Z^{(\mathtt{Uti})}|\mathcal{S}).$

      Alternatively, note that the optimization problem defined above can be reformulated to a min-max problem when $\varphi$ is defined to be the weighted sum as
      \[\min_{(J,\mathcal{S})\in \mathtt{S}^{(\mathtt{dis},\gamma,\alpha_C)}}\max_{(I,\mathcal{R})\in \mathtt{S}^{(\mathtt{emb},\alpha_R,J,\mathcal{S})}} \begin{array}{c}
           \mathbb{H}(Z^{(\mathtt{Pri})}|\mathcal{R}) \\
           + \lambda^{(\mathtt{Uti})} \mathbb{H}(Z^{(\mathtt{Uti})}|\mathcal{S}). 
      \end{array} \]

    \item \emph{Data-based optimality of disentanglement and representation:} Alternatively, the evaluation of a data model and data representation can also be measured independent of a provided objective function. In such case, since the metric is independent of an objective while privacy requirement is imposed in the objective, we assume that our aim is for $\mathcal{S}_{\mathcal{F}}$ to contain as much information as $\mathcal{D}^{(E)}$ as possible while we omit the privacy metric in the discussion. Hence letting $\mathtt{D}$ be the random variable representing the true distribution of the target window, we say that a valid representation model $\mathbf{v}$ is optimal from the perspective of the data if it provides the minimum value for the following optimization problem
    \[\min_{\mathbf{v}\in \mathtt{S}^{(\gamma, \alpha_R,\alpha_C)}}\mathbb{H}(\mathtt{D}|\mathcal{R}).\]
    
    Here we provide an example of such data-based optimality of representation, which is called \emph{reconstructability}. Suppose that $\mathcal{A}$ is an inner product space equipped with a distance function $d:\mathbb{P}(\mathcal{A})\times \mathbb{P}(\mathcal{A})\rightarrow \mathbb{R}$ defined based on the inner product. Here we note that such distance function can be measured based on difference in values or probabilistic measure such as entropy, covariance, or probability expectation. For a value $\epsilon^{(\mathtt{Recon)}>0},$ we say that $\Pi$ is $\epsilon^{(\mathtt{Recon)}}$- reconstructable with respect to the distance function $d$ if for any given any valid data $\mathcal{D}\subseteq \mathcal{A}$ and $\mathtt{E},\mathtt{D}$ the two maps produced by $\Pi.\mathtt{Model},$ we have $d(\mathcal{D},\mathtt{D}(\mathtt{E}(\mathcal{D})))\leq \epsilon^{(\mathtt{Recon)}}.$
    \item \emph{Independence}: Note that in ideal case, the chosen $\mathcal{S}_{\mathcal{F}}=\{Z_1,\cdots, Z_M\}\subseteq \mathcal{R}_{\mathcal{F}}$ needs to be independent. Hence, there needs to be a measure of independence for the chosen random variables. In general, given an independence metric $\psi$ that takes as input a set of random variables and measures the level of dependency between the variables (i.e., the value of $\psi$ increases as the variables depend more on each other), the data modelling process may be optimized based on the minimality of such metric. For any threshold $\kappa>0,$ we say that a model $\mathcal{S}_{\mathcal{F}}$ is $\kappa$-independent if $\psi(\mathcal{S}_{\mathcal{F}})\leq \kappa.$ Here two natural independence metric that can be used are covariance and mutual information. Here covariance between two latent variables measures the \emph{linear} relations between the two latent variables while mutual information measures any relation between them. The use of covariance is much simpler due to its linearity. However, although independent variables must have zero covariance, the converse is not necessarily true. The relation between independence and zero mutual information is stronger where we have equivalence between them. However, it is highly non-linear and hence harder to analyze.
Together with the data-based optimality of representation, we may again define a combining function $\xi:\mathbb{R}^2\rightarrow \mathbb{R}$ that is assumed to be non-decreasing with respect to the first input and non-increasing with respect to the second input.  In such case we say that a valid representation model $\mathbf{v}$ is an optimal representation model from the data perspective if it provides the maximum value for the following optimization problem
    \[\max_{\mathbf{v}\in \mathtt{S}^{(\gamma, \alpha_R,\alpha_C)}}\xi(\mathbb{H}(\mathtt{D}|\mathcal{R}),\psi(\mathcal{S})).\]

    \item \emph{Extrapolation Accuracy:} Suppose that we are given a set of inputs for the protocol, which starts with a valid dataset $\mathcal{D}$ to obtain a data model $\mathcal{S}_{\mathcal{F}}$ along with the two transformation functions $\mathtt{E},\mathtt{D},$ and the extrapolated representation $\mathcal{R}_{\mathtt{Extrapolate}}.$ Here we assume the existence of a statistical distance function $d^{(\mathcal{S})},$ which can be defined, for example through KL divergence or total variance. Let $\mathcal{D}^{(\mathtt{Extrapolate})}\subseteq \mathcal{A}$ be a dataset satisfying the extrapolation query $\mathfrak{q}_{\mathtt{Extract}}$ that optimizes the disentanglement output with respect to the data model $\mathcal{S}_{\mathcal{F}}.$ That is, letting $Z$ be the random variable with the true distribution of the attributes required given the attribute conditions in the extrapolation query, $\mathcal{D}^{(\mathtt{Extrapolate})}$ minimizes $\mathbb{H}(Z|\mathcal{S}_{\mathcal{F}},X)$ for $X\subseteq \mathcal{A}.$
    
    We further let $\mathcal{R}^{\ast}_{\mathtt{Extrapolate}}=\mathtt{E}(\mathcal{D}^{(\mathtt{Extrapolate})}.$ Then for any constant $\epsilon^{(\mathtt{Extrapolate})},$ we say that $\Pi$ provides $\epsilon^{(\mathtt{Extrapolate})}-$extrapolation accuracy with respect to the statistical distance function $d^{(\mathcal{S})}$ if
    $d^{(\mathcal{S})}(\mathcal{R}_{\mathtt{Extrapolate}},\mathcal{R}^\ast_{\mathtt{Extrapolate}})<\epsilon^{(\mathtt{Extrapolate})}.$
    
\end{enumerate}
\section{Detailed Discussion on Components of Data Disentanglement}\label{sec:Discussion}

\subsection{Data Extraction}\label{sec:Discussion-DataExtraction}
Given the input data, the requested data may only involve a small portion of the data, characterized by the extraction query. Here, we discuss extraction in two directions: record extraction and attribute extraction.

Firstly, to support the overall motivation of learning the true distribution for the objective function defined over the data satisfying the query, the given data sample, regardless of its size, can never provide full information regarding the true distribution. Considering more data records sampled from the desired target window may improve our confidence in our distribution estimation while including data records sampled from a different distribution may instead reduce such confidence. Hence the task of record extraction can be reduced to the task of identifying data records that provide information regarding the target window while removing data records that provide more noise than information regarding the target window. 

Here trivial solutions that take either just the target window or the whole input data as input for the next step may not be the optimal solution. On one hand, taking only the target window restricts the information that may be learned regarding the desired true distribution. On the other hand, taking the whole input data directly prevents us to learn the actual distribution that we are interested in since the resulting learned model will model the overall structure of the whole data instead of the desired target window. Although it can be argued that given the overall structure, we may also get information about the structure of the target window, due to the limited resources available for the learning process, it implies that the amount of information we may have regarding the target window will be smaller. We further note that even if we still focus on the learning of the target window from the whole data, including all data implies that we may also include data with small relevance to the training data. In such approach, data with low relevance to the target window may introduce more noise than information that we may benefit from, which may cause the overall information we may learn about the target window worse than before. 

Apart from extraction of records based on their relevance to the target window, another extraction to be considered is the attribute extraction. Here, we note that, again, due to the interrelationships among attributes, by including attributes with strong relationships with the target attributes, the amount of information the model can learn may increase. Due to the potential additional information, it is then desired to select attributes that may contribute the most amount of additional information regarding the attributes targeted by the extraction query while omitting those with minimal amount of additional information on the target atrributes. We again note that a trivial solution of just including the target attribute to the extracted data may prevent the model to learn from any other attributes. On the other hand, including all attributes in the extracted data enables the model to learn any possible interrelationships that the target attributes have. However, without careful design of how the model chooses the interrelations to be learned, it is possible for it to choose a stronger relation irrelevant to the target attributes.

A natural question regarding such optimization is the design of evaluation metric to judge the performance of an extracted data. Here for simplicity, here we first consider the design of evaluation metric when the request does not come with any objective function. In such case, our objective is to learn the true distribution of the target attributes under the given extraction attributes condition. We note that the discussion below can be easily extended to the case when the request comes with an objective function. Furthermore, such evaluation function described here is only considered for illustration and a more careful design of such objective function can be seen as one of the main challenges in data disentanglement in general.

Here, for our illustration, we consider a simple structure where given the extraction query, the extracted data is determined and directly used as an input to a publicly available disentanglement protocol from TabSyn \cite{tabsyn}. We then evaluate the generated set of latent variables. Here we consider two evaluation functions, one for the record selection and another for the attribute selection.

Firstly, we consider an evaluation function for the extracted records while keeping the list of attributes the same. Before we discuss our evaluation function, recall that here, regardless of the number of records being used, we are not aiming to learn the overall distribution of the sample nor the sample distribution of the extracted records. Instead, we are interested in learning the true distribution of the random variable representing the data satisfying the extraction attribute condition. A simple way to evaluate the performance of the extracted record is through its average capability to predict a target attribute's values under the attribute condition requirement. This can be done by first taking some records from the input data that satisfies the condition as the test data. The extraction process is then used to generate the extracted data from the remaining data records. We then use the extracted data to train a classifier to predict a target attribute which is randomly chosen from the attributes in the target window. We evaluate the classifier using 2 common classification scores, the Average Precision and the ROC-AUC scores. Here we use such evaluation metrics to test the two trivial solutions as well as a simple record extraction protocol. 


A simple record extraction protocol can be achieved via classification-based evaluation. A classification-based approach attempts to identify additional records that has sufficiently high probability to be sampled from the target true distribution. More specifically, in order to help us in identifying more records that should be added to our extracted data, we adopt the approach proposed in Positive-Unlabeled (PU) learning framework \cite{bekker2020learning}. Such approach starts by defining a new attribute which represents whether a record belongs to the target window. We label the records in the target window as ``positive'' while records outside the target window, which we will call candidate records, to be labeled as ``unlabeled''. The PU approach consists of multiple iterations where each iteration is used to refine a classifier outputting the probability of each candidate record being in the target window, that is, the probability of each candidate record to have ``positive'' as its value in the added attribute. Here, we note that a classifier training needs both positive and negative labels. To achieve this, for the first iteration,  a random subset of the unlabeled candidate records are assigned negative labels. The pair of positive and negative labelled sets can then be used to train a classifier for such label. This classifier can be used on the remaining unlabeled candidate records to assign a probability value that such record has a positive label. Based on these probabilities, using a predefined threshold, two larger sets can be defined with one set containing all records with either positive label or sufficiently high probability output while another set containing all records with either negative label or sufficiently low probability output. Here having the two new sets, it can then be used as training data for the next iteration's classifier. Here this can be repeated multiple times to obtain an overall classifier providing probabilities for candidate records to belong to the target window.

In our experiment, such approach is done for $100$ iterations. The classifier trained in the last iteration is then used to generate the probabilities of each candidate records to be from the target window. Along with a predefined threshold, we can then extract records having sufficiently high probability as the extracted data. 

Note that in general, the effect of including a data record in terms of the amount of knowledge we gain regarding the true distribution is dependent on the attributes being considered. In our experiment, we consider two possible sets of attributes where we consider a small target window as a baseline $\mathcal{C}_1$ and a larger target window where we also include other relevant attributes $\mathcal{C}_2.$ A more detailed discussion on how such relevant attributes are chosen can be found in the next part regarding attribute extraction. The experimental result can be observed in Table \ref{tab:rowres}.
\begin{table} 
    \centering
    \caption{ML Efficacy results of our simple extraction protocol}
    \begin{tabular}{|p{2cm}||p{2cm}|p{2cm}|}
        \hline
          & Average Precision Score & ROC-AUC Score \\
          \hline
          \multicolumn{3}{|c|}{\textit{Initial Target Window $\mathcal{C}_1$ (baseline)}}\\
          \hline
         Original records& \textit{11.36\%} & \textit{81.82\%} \\
         \hline
         After record extraction & 9.31\% & 79.77\%\\
         \hline\hline
         \multicolumn{3}{|c|}{\textit{Enlarged Target Window $\mathcal{C}_2$}}\\
         \hline
         Original records & 19.56\% & 91.20\% \\ 
         \hline
         After record extraction & \textbf{20.78\%} & \textbf{91.70\%} \\ 
         \hline
         \hline
    \end{tabular}
    \label{tab:rowres}
\end{table}


As can be seen from the experiment, the additional records have minimal or even detrimental effect to the ML efficacy figure in the initial target window.  This may suggest that in the original target window, relation between attributes is not as strong. Hence, instead of having the extra record as supporting evidence on such relation, it may provide noise causing the decrease in the Average Precision of the resulting classification model from $11.36\%$ to $9.31\%.$ 

On the other hand, when more attributes are introduced which provide stronger interdependencies between attribute, such extra records provide stronger evidence on such information, allowing the disentanglement method to learn more effectively, exhibited by the increase in the ML efficacy evaluation figures. This supports our claim that the effectiveness of including more records depend on the existence of interdependencies between attributes for the extra records to serve as supporting evidence.

This result further motivates the need of attribute selection to be done for the extraction data. Furthermore, it also motivates the need that the two selections are done together instead of only having one of them.

\begin{table} 
    \centering
    \caption{Average Mutual Information of latent variables on target window's attributes}
    \begin{tabular}{|p{3cm}||p{3cm}|}
        \hline
          Data Modelling Input Attributes & Mutual Information score\\
          \hline
         Original target attributes (5 Attributes) & 3.9653  \\
         \hline
         After Simple Attribute selection method (8 attributes) & 4.0124 \\
         \hline
        Including all Available Attributes (15 attributes) & 4.0273 \\
         \hline
    \end{tabular}
    \label{tab:colres}
\end{table}

Secondly, we consider an evaluation function for the extracted attributes while keeping the list of records the same. Here we design the evaluation function based on the observation that a good set of latent variables should have large mutual information with the set of target attributes. Hence, we evaluate the set of latent variables by the average mutual information score between the latent variables and the target attributes. Along with the evaluation metric we used in the record extraction step, we would like show the possibilities of such evaluation metric to be either information theoretic based or downstream task based. Furthermore, such evaluation can also be done either locally for each component or globally for the whole disentanglement framework.

We consider the two trivial solutions of either using just the target attributes (5 attributes) or all available attributes (15 attributes). We compare such trivial solutions with the simple way of column selection by including attributes with the highest correlation with the target attributes. 
Here for each possible set of attributes used for the input of the data modelling and model analysis, we measure the average information regarding the target attributes that can be learned from the resulting latent variables. Intuitively, since adding attributes will always increase the amount of information that is available regarding the target attribute, we can expect that such evaluation metric will have the lowest value when we only consider the target attributes while having the highest value when we consider all available $15$ attributes. As observed in Table \ref{tab:colres}, such intuition is indeed confirmed where we have the increase in average mutual information when considering all attributes. Here, we note that in our simple attribute selection process, instead of requiring all the $10$ additional attributes, we only consider the inclusion of $3$ additional attributes, i.e., $30\%$ of the cost. However, as can be observed in Table \ref{tab:colres}, such minimal amount of addition, amounts to up to $76\%$ of the maximum possible gain obtained by including all $10$ additional attributes. This demonstrates the importance of an attribute selection protocol that balances the computational efficiency and the performance improvement.


We note that in either case, such selection may be influenced by external knowledge provided in the input. For instance, a known functional dependency between target attributes and other attributes may be used to determine the inclusion of such attributes. Knowledge about the true distribution of the target window may also help in determining the records to be extracted, which helps in gaining more knowledge in the true distribution of the target window.

Next, we briefly review some existing techniques of extraction that have been considered.  We would like to remark that to the best of our knowledge, the existing data extraction mechanisms are not exactly designed to address the same problem as the one we have previously discussed. Firstly, in general, the aim for such data extraction solutions has been more focused on the extraction of data with similar behavior as the original input data. Hence, in contrast to our direction that we aim to learn the true distribution of the target window, which may not be the same as the sample distribution given by the sample target window, such works assume that the sample distribution is the target distribution to be learned and the extraction is done to generate a different dataset that exhibits a similar behaviour as the original data sample. Secondly, there is also a difference in the format of the input data. Recall that in our framework, we have a large dataset and a smaller subset of such dataset. The aim is to generate a different subset of the large dataset to help in learning the target distribution. As will be discussed later, in existing solutions, we either only have the larger dataset where we want to generate a representative smaller subset or we only have the smaller dataset where we want to generate more synthetic data to enrich the dataset while exhibiting similar behavior.

However, reviews of such solutions may provide us with some insights on the development and improvement of the data extraction component in our framework. In general, existing data extraction solutions attempt to address one of the following two problems. Firstly, suppose that we are given a large data source. In such case, it may not be feasible to process the whole data due to the resource limitation. In such case, it may be desirable for us to generate a smaller subset of the data that is representative of the whole data, which can then be processed as needed. Using the smaller subset proves to be beneficial due to its cost effectiveness \cite{datasetdistillation,dsdm,lessismore,datasetdistillationsurvey}.

On the other hand, data extraction may also be used when the training data is too small. In such case, we may not be able to get a good model of the data due to the insufficient information. In such case, some works try to enrich the data source by generating a synthetic data using a simpler technique that has similar behavior as the original training data. Such works include SMOTE \cite{smote} and Massive Trend Diffusion \cite{generativemtd}.

Here there have also been works discussing on how such extracted data should be evaluated. In general, since our aim is to have such extracted data to exhibit a similar behavior of the original input data, evaluation is done based on the similarity of the output and the input data. One popular way of measuring such similarity is thorugh the similarity of its performance. More specifically, here the two dataset can be used to train two different predictive models. Then it is argued that the extracted data is evaluated based on the accuracy difference between the two predictive models \cite{matchingtrainingtrajectories}. On the other hand, a different way of measuring the similarity between the two sets can simply be from their statistical properties themselves. In this case, performance of an extracted data can be measured based on the statistical difference between the distributions of the two datasets \cite{dream,distributionmatching}.

Lastly, we briefly discuss some challenges that may be essential in the study of extraction process.
Firstly, the design of evaluation metric both locally in the extraction step and globally for the disentanglement protocol in general. Such evaluation metric can be used in both evaluating the extracted data and optimizing such extraction process. Intuitively, since the overall objective of data disentanglement is to learn the true distribution of the target window, a natural approach is through the conditional entropy of the target random variable given the knowledge from the extracted data. We also note that an evaluation metric based on a specific downstream task can also be considered. In such case, the performance of the extracted data is measured by the performance of, for instance, the accuracy of the classifier trained based on such extracted data. We note, however, that such approach may be more appropriate for a more global evaluation metric for the whole disentanglement protocol and less suitable for the more specialized evaluation where only the extraction process is considered. As illustration, our experiments above provided two possible forms of evaluation functions. The first evaluation function is a global function which directly evaluates the disentanglement protocol. On the other hand, the second evaluation function is a local function only evaluating the column extraction. However, it is done using a specific downstream-task for evaluation, namely, classification of target attributes.

Furthermore, it is natural for the extraction process to rely on its performance in the latter stage of the disentanglement. However, such disentanglement process cannot be conducted without having the extracted data as an input for the training step. Firstly, this suggests that an optimal way of designing the extraction process should not be independent of the disentanglement process, justifying the need to include such extraction process in our disentanglement framework instead of having it as a preprocessing phase independent of the framework. Secondly, this may provide a challenge in designing a way to estimate the expected performance of the latter stage of the disentanglement process given an extracted data, which may be used in the optimization process. Alternatively, this suggests that a disentanglement solution that contains extraction step may require the extraction process to work closely with the other steps to approximate the optimal extraction as well as optimal data representation.

\subsection{Data Modelling}\label{sec:Discussion-DataModelling}
The next component to be discussed is the data modelling. We note that data modelling, along with the next component, model analysis, are the focus of most disentanglement solution. In this section, we focus in the former, data modelling. 

Having the extracted data, data modelling attempts to identify the structure of the extracted data, which will be essential in learning the true distribution of the desired random variables corresponding to the target window. More specifically, the aim of data modelling is to construct a set of random variables which are functions of the attributes in the extracted attributes such that such random variables contain sufficient information on the target variable.  Here we call such set of random variables the set of latent variables. We also require that it also comes with an approximation of the inverse of such functions to map values in latent space back to the extracted attributes. Here, following the term used in a family of disentanglement solutions called autoencoder (See, for example \cite{VAE,switchtab,ContrastiveVAE}, we call the function sending data from real space of extracted attribute to the latent space and the inverse as the encoder and decoder corresponding to the data model.

We further note that in the effort of choosing the set of latent variables, a protocol comes with a pre-defined relationship family which serves as the search space for such latent variables. Here solutions with a deterministically determined set of latent variables can be seen to have small relationship family where selection is done deterministically. This includes approaches such as factor analysis \cite{factoranalysistutorial}\cite{oivae}.

There are in general two main objectives of data modelling which determines the latent variables to be selected. Firstly, as has been mentioned, such set of latent variables need to contain sufficient information on the target random variables. This allows the model to generate a good representation of the extracted data in the latent space which is essential in the understanding of the extracted data and the objective random variable. To achieve this, in general, a latent variable can be used in two different ways. Firstly, it may be used to describe an overall behavior of the data. For instance, a latent variable may represent a relation between some attributes which can be observed throughout the extracted data. Such variable is generally called the common feature. Alternatively, it may instead represent a relation which can be used to provide contrast between several subsets of the extracted data, which is generally called unique feature. In such case, such relation may provide distinguishable behavior in the subsets of extracted data. Such relation can then be used to provide a more detailed variations of the model following the different subsets of the data. Such approach can be observed, for instance in \cite{switchtab,ContrastiveVAE,TripleDisentangled,DivergenceDisentanglement,UNTIE}. To accomodate unique features, we impose that the latent variable should also come with an auxiliary output listing the different subsets of data records where separate analysis of the latent variable in the different subsets may provide us with the contrasting description which may help us in representing the data. This necessitates the auxiliary output we assumed the data model has. For common features, such auxiliary output will only contain one set containing all the extracted records. 

The second main objective of data modelling is regarding the statistical dependence of the latent variables. Ideally, we would like the chosen set to be statistically independent. However, it may not be feasible to require such a strong condition. So instead, the aim for data modelling is to weaken such statistical dependence between the latent variables, which can be seen as a justification of the term ``disentanglement''. This is due to several reasons. 

Firstly, note that having such set of latent variables is not sufficient in fully understanding the extracted data. We need to analyze such variables further, which is the main objective of the next component, model analysis. For instance, such analysis may involve the estimation of the distribution of such latent variables. However, such analysis may not be feasible when the set of latent variables is not independent. For illustration, the estimation of the distribution of a set of non-independent latent variables requires the estimation of the joint distribution. However, when joint distribution is considered, the sample space is the direct product of the sample space of the sample space of each latent variable. This means that the size increases exponentially with respect to the number of latent variables. Since the accuracy of such estimation depends on the size of the sample, when the sample space grows exponentially, the required size of such sample will also grow exponentially. So in most cases, joint distribution estimation is not feasible. 

Secondly, we recall that although such model can be used to describe the original extracted data, we may also use it to investigate how the data may react for different external stimulus. Note that if there is a strong dependence between the latent variables, such external stimulus must affect some, if not all, latent variables. This makes such analysis much more complex. In contrast, when the latent variables are independent, changing one of the latent variables will not have effect on the other latent variables. This allows us for more precise analysis of the effect of different external stimulus.

As previously described, existing solutions work under a relaxed requirement that the latent variables have less statistical dependence. Note that due to such reduction of statistical dependence, we may feed the output of a data model to another data modelling to provide a new set of latent variables with even less statistical dependence. Such approach is called stacking \cite{HQVAE,NVAE}. Another approach is to let such statistical dependence remain. However, the protocol is then equipped with an additional subprotocol to either learn such dependencies or reducing it further. Towards the first direction, some works aim to define the dependencies explicitly \cite{CausalVAE, CausalDisentangled}. By having an explicit model of the dependencies, the interaction between the latent variables can be predicted, which can further simplify the analysis of the latent variables. Alternatively, the data modelling step itself can also be modified to enforce the independence requirement further. This is done by modifying the loss function where a set of latent variables will be penalized based on different independence metric \cite{factorvae,betavae}.

Similar to data extraction, the process of data modelling may benefit from the external knowledge. A functional dependency may either be used to fix one of the latent variables to represent such dependency or, if we deem such relation may be imposed in the post-processing, we may remove such relation from the relationship family to ensure that such relation is not chosen. The determination of whether a latent variable represents a unique feature may also be done by the help of the external knowledge of the distribution of some attributes. 

It is easy to see that since data modelling is one of the two main components where the main objective of data disentanglement is carried out, its inclusion in the data disentanglement process is compulsory.

Next, we briefly discuss some possible challenges that need to be considered in the investigation of data modelling. Similar to what has been discussed in the previous section, there needs to be a rigorous discussion on an evaluation function that can be used to estimate the performance of a set of latent variables. Here we again can have such evaluation function to measure both local performance of the data modelling or the global performance of the data disentanglement as a whole. It can also be designed to directly measure the performance or be based on a specific downstream task, such as classification performance. Secondly, in general, machine learning based solutions such as VAE or GAN \cite{VAE,ctgan} determines such latent variables including the corresponding encoder and decoder without having explicitness of the relation between a latent variable and the extracted attributes. On one hand, having non-explicit relations in the relationship family allows the relationship family to be larger, providing a higher confidence in the performance of the data model. However, a non-explicit latent variable may cause further analysis or modification of such latent variable to be difficult. One of the main challenges that data modelling has is to design a protocol with comparable relationship family with existing solutions without having any non-explicit relations in the family. Next, to the best of our knowledge, most disentanglement solution aims to learn the true overall distribution of the extracted data. This can be seen, for instance, in solutions built in the direction of autoencoders \cite{VAE,tabsyn,factorvae}. In such solutions, the loss function contains reconstruction loss, which evaluates a solution based on its capability to reconstruct the original extracted data. However, such reconstructability may not be the priority for any objective function. The challenge of designing an objective-based disentanglement is both interesting and relevant to its real life use where in most cases, request is made not to reproduce the original data. Instead, the requester may want to use such information to, for instance, create a classifier to help in their decision making process. 

The last challenge we want to discuss is regarding imposing the independence requirement in the data modelling process. For instance, the loss function of AE-based solutions \cite{VAE, betavae, tabsyn} only penalizes a solution based on the \emph{covariance} of the latent variables, which only measures the linear statistical independence between the latent variables. Here, designing a solution that reduces the overall statistical independence is a challenge that may be worth investigating. Another possible direction is through the use of deterministic post-processing of the latent variables. More specifically, here, instead of designing a data modelling solution that directly outputs a statistically independent set of latent variables, we may have an additional protocol which, given the set of latent variables, outputs a new set of latent variables with less or no statistical dependency while still providing approximately the same amount of information as the original set of latent variables.

\subsection{Model Analysis}\label{sec:Discussion-ModelAnalysis}
As has been previously discussed, model analysis is the second half of the focus for most of the disentanglement solutions. In model analysis, given the model produced by the data modelling component, the model analysis component aims to provide a deeper analysis of the latent variables in the subsets of the extracted data provided as auxiliary output of the data modelling. Here in general, the model analysis aims to produce an estimation of the distribution for the latent variables in such subsets. Typically, such analysis starts by assuming that each latent variable is statistically independent, allowing the distribution estimation to be done for each latent variable separately. 

For each latent variable and each subset assigned to it, our aim is then to estimate the distribution of such latent variable given the sample provided by the subset of the extracted data. The distribution estimation has many different possible approaches. The most common way of doing such estimation is through the use of parameter estimation. Here, the latent variable is first assumed to follow a specific distribution, for instance, normal distribution. Having this, to estimate the distribution, it is sufficient to estimate the parameter defining such distribution. So, for instance, when the distribution is assumed to be normal, it is sufficient to estimate the mean and the variance of each of the latent variables. Here one of the challenges is that the sample may be noisy which presents a further challenge in the distribution estimation. A solution such as the diffusion model \cite{tabsyn} can then be used to identify and remove the sample noise and produce the estimated distribution. Note that such approach works particularly well in the work of disentanglement for image data \cite{factorvae}\cite{betavae}. However, it may not be as effective when considering tabular data where data distribution may follow different distributions. In such case, treating the sample to follow a more complex distribution may help in obtaining a more accurate distribution. For instance, the sample may be considered to follow a mixture of Gaussian distribution. So in this case, solutions using Gaussian mixture model \cite{VAEGMM, GMVAE, tGMVAE} generates a further partition to the data sample where each subset follows different Gaussian distribution. Note that approaches done in \cite{GMVAE, tGMVAE} take advantage of the Gaussian mixture model in establishing the subsets that should be output along with the data model instead of the distribution estimation for each of such subsets. 

Alternatively, a distribution estimation can also be done without first assuming the overall distribution structure. Such approach is called the non-parametric approach, which can be achieved, for example, using the kernel density estimators \cite{kdeapplications}\cite{kdeeconomy}. We note that although such approach may be more general without any need of an assumption for the overall structure of the dimension, it is in general less accurate and when the dimension of the sample space is large, it becomes infeasible to do so.

We again note the benefit of external knowledge in such model analysis. Firstly, any distribution requires the definition of the sample space. Having such knowledge enables the distribution estimation to be done without the need of the estimation of such sample space, which may not always be accurate, especially if some possible values are not present in the given data sample. Secondly, if the distribution of a specific latent variable is known, then parametric approach can directly be used without the need of further assumption. 

Furthermore, note that, in general, the data extraction, data modelling, and model analysis components need to work simultaneously together since the optimization in one component depends on the output of the other. In our work, we specifically separate them in terms of functionality while also allowing them to work together to improve the performance of each other. Note that, however, it is clear that model extrapolation should be done in a latter stage due to the different overall objective. Here in contrast to the other three components of which the objective is to learn from the input data, the objective of model extrapolation is to predict the latent representation in a potentially different scenario which can be absent from the original data. However, this, again, does not mean that during the model extrapolation step, other components cannot be involved. In some cases, returning back to the existing data allows for information that was not previously learned to again be included to provide a more accurate extrapolated representation.

We briefly discuss some possible challenges on the model analysis component. Firstly, as has been previously discussed, one of the main challenges is the design of an evaluation function to estimate the performance of the model analysis component. Recall that our aim is to estimate the true distribution of the target window or the random variables related to such target window. However, since such true distribution is unknown, there needs to be a way to estimate the expected error of the estimate produced by the model analysis. 

Secondly, note that, in contrast to works on image data, distribution of latent variables may not always be approximated by normal distribution. In such case, if the distribution is not included in the external knowledge, there needs to be a way to perform the model analysis. As has been previously discussed, if the overall structure is unknown, non-parametric approaches like kernel density estimation may be used but it may not be suitable, for instance, if the size of the sample space is too large. A different approach may be interesting to consider.

Next, we recall that although statistically independent latent variables are ideally desired, it may only be possible to have an approximation towards such independence. In this case, treating the latent variables to be independent and separately estimate the distribution may cause further error in our estimation. A natural challenge to be considered is hence the distribution estimation when we consider the weak statistical dependence between the latent variables into account. A possible approach that may be considered is through the use of approaches like the copula-based approach. This approach allows the marginal distributions of each latent variables to be considered separately while the copula is used to model the dependency between such latent variables. We note that although such approach works for any possible set of latent variables, due to its generality, we may not take advantage of the fact that the data modelling aims to reduce such statistical independence. Hence it may be interesting if we can gain a better approach which takes advantage of the weak statistical dependence.

\subsection{Model Extrapolation}\label{sec:Discussion-Extrapolation}

Lastly, given the data model as well as the data representation output by the data modelling and model analysis given the extracted data as an input respectively, such data representation provides us with information regarding the target window distribution and behaviour. Although such information is important in the understanding of the input data, it may not be the aim of the request. A natural aim that is common in the real life application of disentanglement that needs to be considered is the use of such information to estimate the data representation for different situation. For instance, a data model for fraud-related data in a country for a specific time range may be used to predict the fraud behaviour in said country for a future time range. Alternatively, it may also be used to predict the effect of a similar fraud in a different country. 

In such case, the component encodes the extrapolation condition and uses it to generate a new data representation which represents the hypothetical data that we would have received in such hypothetical scenario. We note that such component is in general less explored. In the field of image disentanglement, disentanglement is conditioned on a label, which is a text string assigned to the image. A conditional extrapolation based on input to generate images with characteristic following the description provided by the label has been investigated in \cite{ccvae}. On the other hand, the study of conditional generation is explored in a specific scenario where continuous attributes are generated conditioned on a discrete attribute. Such solution is build based on generative adversarial network method \cite{ctgan}. Although studies of conditional generation where the condition is a discrete attribute has also been considered in VAE architecture \cite{dptvae}, in general the problem of conditional extrapolation and data generation has not been well explored.


We note that although such component may not be essential in the learning process for disentanglement, it becomes essential in how the learning result may be used. Here the extrapolation condition includes simple conditions such as the different data distribution. For instance, here, we may require the representation to consider the scenario where the gender distribution is different from what the data has. Alternatively, if we are more interested, for instance, in learning the fraud cases instead of its distribution among all other data, requiring the representation to have a larger proportion of fraud may be useful. On the other hand, complex conditions such as those requiring us to extrapolate the data from different country may involve different overall distribution of various variables. In any of such case, the aim of model extrapolation is to translate such conditions to an explicit transformation of the data representation. 

Furthermore, we note that although model extrapolation may not help in the learning process, it does not mean that other components may not be useful in the model extrapolation. For instance, after the model has been established, it may make sense to have a deeper investigation of the training data using the former three components to provide more information that is essential in the extrapolation process. Due to this relation, by treating extrapolation as a separate post-processing step may cause the loss of information that the model could have learned otherwise.

Lastly, we discuss some potential challenges that the model extrapolation component has. Firstly, as has been previously discussed, a more rigorous study on the design of evaluation metric is essential. Note that evaluation metric that may apply in the model analysis component may not be directly applicable in this component. More specifically, an accuracy metric that measures the amount of information regarding the data is stored in the latent variables may not be useful since we may not have sufficient amount of data in the input data that is relevant in the desired hypothetical situation. Because of this, a different accuracy evaluation function may need to be designed.

Secondly, as has been discussed, here the component attempts to transform the given extrapolation query to a transformation function of the latent variables. Such transformation may not be trivial and a deeper analysis is required to design how such component can be realized. This challenge is further compounded with the fact that in the most state-of-the-art disentanglement process, the actual transformation from attribute space to latent space and vice versa is non-explicit and only approximated. In such case, it becomes more difficult to translate the hypothetical situation to the latent space. Similar to the data modelling component, when an objective function is provided, such transformation should also take such objective into account. In such case, the transformation can be more prioritized towards the conditions that are more relevant to the objective given. 
\end{document}